\documentclass{article}

\usepackage[utf8]{inputenc} 
\usepackage[T1]{fontenc}    
\usepackage{hyperref}       
\usepackage{url}            
\usepackage{booktabs}       
\usepackage{amsfonts}       
\usepackage{nicefrac}       
\usepackage{microtype}      
\usepackage{lipsum}

\usepackage{nips14submit_e,times}

\usepackage{graphicx}
\usepackage{float}
\usepackage{amsmath}
\usepackage{amssymb}
\usepackage{subcaption}

\graphicspath{./images/}

\newcommand{\projectTitle}{Context Encoding Chest X-rays}

\def\chestxray{\textit{ChestX-ray14} }

\title{\projectTitle}

\author{
  Davide Belli$\quad \quad \quad \quad$ \\ 
  University of Amsterdam$\quad \quad\quad \quad$\\
  \texttt{d.belli@uva.nl}$\quad \quad \quad \quad$ \\
  \And
  Shi Hu$\quad \quad$ \\
  University of Amsterdam$\quad \quad$ \\
  \texttt{s.hu@uva.nl}$\quad \quad$ \\
  \AND
  Ecem Sogancioglu \\
  Radboud University Medical Center \\
  \texttt{ecem.lago@radboudumc.nl}
  \And
  Bram van Ginneken \\
  Radboud University Medical Center \\
  \texttt{bram.vanginneken@radboudumc.nl} 
}

\nipsfinalcopy

\begin{document}
\maketitle

\begin{abstract}
Chest X-rays are one of the most commonly used technologies for medical diagnosis. Many deep learning models have been proposed to improve and automate the abnormality detection task on this type of data. In this paper, we propose a different approach based on image inpainting under adversarial training first introduced by \cite{gan}. We configure the context encoder model for this task and train it over 1.1M 128x128 images from healthy X-rays. The goal of our model is to reconstruct the missing central 64x64 patch. Once the model has learned how to inpaint healthy tissue, we test its performance on images with and without abnormalities. We discuss and motivate our results considering PSNR, MSE and SSIM scores as evaluation metrics. In addition, we conduct a 2AFC observer study showing that in half of the times an expert is unable to distinguish real images from the ones reconstructed using our model. By computing and visualizing the pixel-wise difference between the source and the reconstructed images, we can highlight abnormalities to simplify further detection and classification tasks.
\end{abstract}


\section{Introduction}
\label{introduction}

Detection of abnormalities in medical images is a key step in diagnosis, monitoring and treatment of diseases. Among different detection techniques, X-ray scanning is a cost-effective and frequently used solution to inspect bone structures and diseased tissues. Chest X-rays in particular capture the major organs in the body and enable doctors to effectively examine patients.
Many CAD (Computer Aided Diagnosis) systems have been developed to help doctors in this complex task given the critical importance of an accurate detection. Example of diseases where automated detection with this technologies is already available are diabetes \cite{diabetes} and breast cancer \cite{icad}. Following recent advancements in the artificial intelligence field, researchers have been proposing various deep learning models to automate the X-ray detection problem. Contributing towards this goal, the National Institutes of Health publicly released in 2017 the \chestxray dataset containing over 110,000 labeled images.

In this project, we try an unsupervised approach to chest abnormality detection using deep learning models under generative adversarial training. We show that these models learn to produce a compressed latent representation of features such as shapes and textures in the healthy tissues and bone structures of scans. This representation is then used to generate an healthy version of the original chest X-rays of a patient, which can be compared to the original one to recognize possible abnormalities. The shape of the abnormality can be obtained by performing pixel-wise difference between the patches. This abnormality detection process could also be automated and used to help doctors in diagnosis. In this project, we use the context encoder under adversarial training and focus on finding the best model configuration to generate realistic healthy X-ray chest patches.

\subsection{Related Work}

The main goal of this work is to successfully detect abnormalities in medical images.
Among the various advancements in Computer Vision for image analysis, Convolutional Neural Networks \cite{cnn} have been proven to be particularly effective for detection, classification and segmentation tasks due to their space and shift invariance and the little need of pre-processing.
Recently proposed R-CNN architectures like \cite{rcnn, fast-rcnn, faster-rcnn} and in particular Mask R-CNN \cite{mask-rcnn} further improve results in segmentation tasks by building on top of the basic CNN.
For classification tasks, \cite{resnet} and \cite{inception} are frequently used state-of-the-art models.

\subsubsection{Generative Training} 
This work, however, aims to solve the detection problem in an indirect way. Instead of trying to detect abnormalities in the original X-ray scan, we want to apply generative models to reconstruct an healthy version of the X-ray scan, which can be compared to the original one to find possible abnormalities.
Generative models have been widely applied in recent research advancements, starting with GANs \cite{gan} and then incorporating the convolutional layer in the DCGAN \cite{dcgan} architecture. 
An impressing number of follow-up works showed how training using the generative setting can be applied to perform a multitude of tasks on image datasets.
For instance, very recently \cite{imagetoimage} proposed \textit{pix2pix}, a version of the conditional adversarial network based on an encoder-decoder architecture. Their model is able to accomplish a variety of new tasks, like recoloring black and white images, generating maps from aerial pictures, street scenes from labels and objects from sketches. This demonstrates how generative training can be applied to a variety of new tasks in addition to the more common ones like classification and segmentation.

\subsubsection{Self-supervised Learning and Image Inpainting} 
Particularly meaningful for our goal is the work from \cite{contextprediction}, which turns unsupervised training into self-supervised. The authors use the spatial context around a central patch as a signal for training over a localization task, where their goal is to learn a meaningful feature representation of the patches. The main idea is that by compressing visual features in a low-dimensional embedding vector, a rich representation is necessary to make sense of the content and correctly complete the prediction task. Pathak \textit{et al} \cite{context-encoder} further expand this self-supervised approach to the image inpainting task which we are going to consider in our work. The context encoder they propose aims to inpaint the missing central patch of an image given the surrounding context. Using an encoder-decoder architecture, they ensure that the information contained in the context area is compressed into an low-dimensional vector. Finally, they apply generative adversarial training to teach the decoder how to reconstruct the central patch given the representation of the context. The results they show demonstrate that the context encoder improves on previous architectures when applied to ImageNet and StreetView images. In our work, we explore how the context encoder architecture can be applied to high resolution \chestxray images and how to modify the configuration and the model itself to obtain better results.

Other solutions for image inpainting using generative training have also been introduced recently \cite{semantic-inpainting, contextual-attention}. 
In a project connected to this work \cite{inpainting-shi} we compare the inpainting results on \chestxray described here with the ones obtained using other aforementioned solutions.

\section{Dataset and Methodology}
\label{dataset}
The \chestxray dataset collects 112,120 frontal-view X-ray images from 30,805 unique patients. The dataset is extracted from the clinical PACS database at National Institutes of Health Clinical Center, considering around 60\% of all frontal X-rays. For this reason, \chestxray dataset should accurately describe the true distribution of diseases and clinical reports in the population. 
The labels provided with the dataset describe what type of abnormalities are present (if any) in every chest X-ray. The categories describing the pathologies are: \textit{Atelectasis, Consolidation, Infiltration, Pneumothorax, Edema, Emphysema, Fibrosis, Effusion, Pneumonia, Pleural Thickening, Cardiomegaly, Nodule, Mass} and \textit{Hernia}.

\begin{figure}[h]
    \centering
    \begin{subfigure}{0.3\textwidth}
        \centering
        \includegraphics[width=\textwidth]{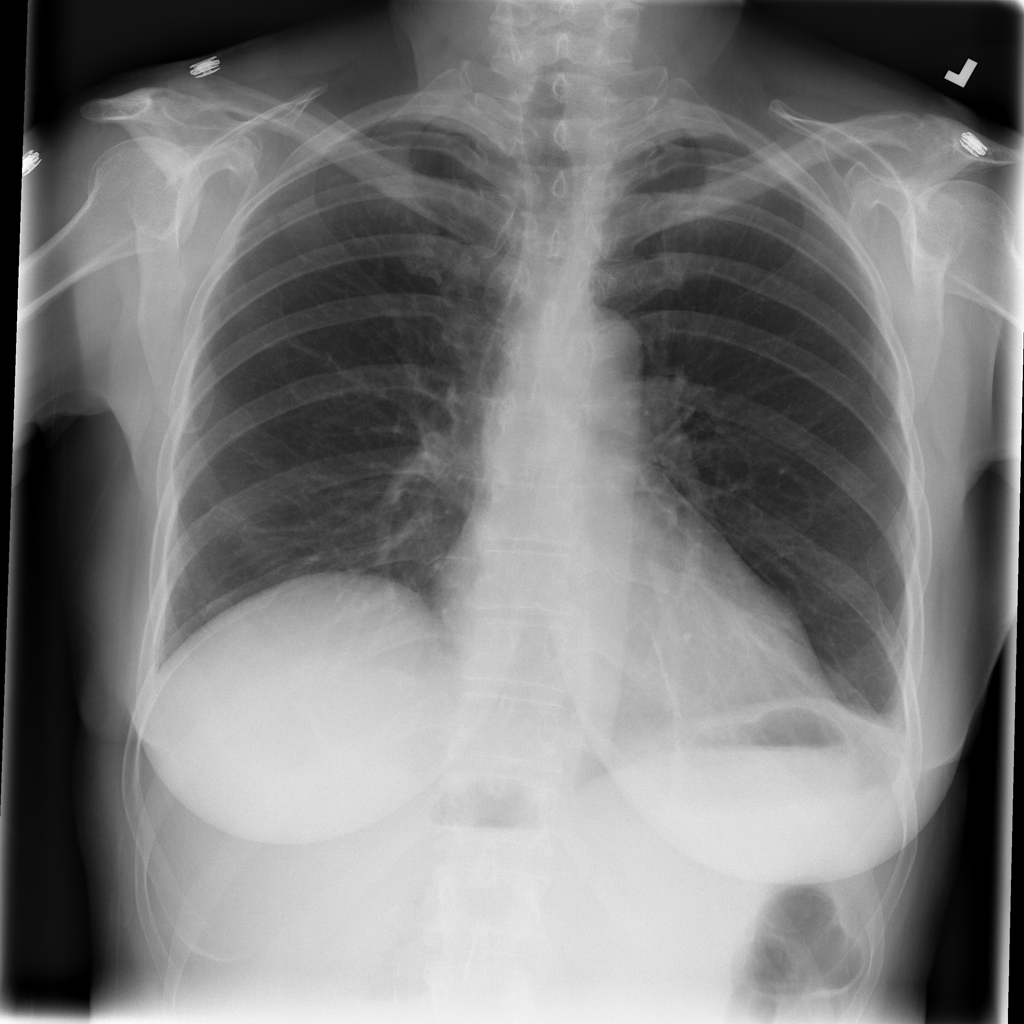}
    \end{subfigure}
    \hspace{5mm}
    \begin{subfigure}{0.3\textwidth}
        \centering
        \includegraphics[width=\textwidth]{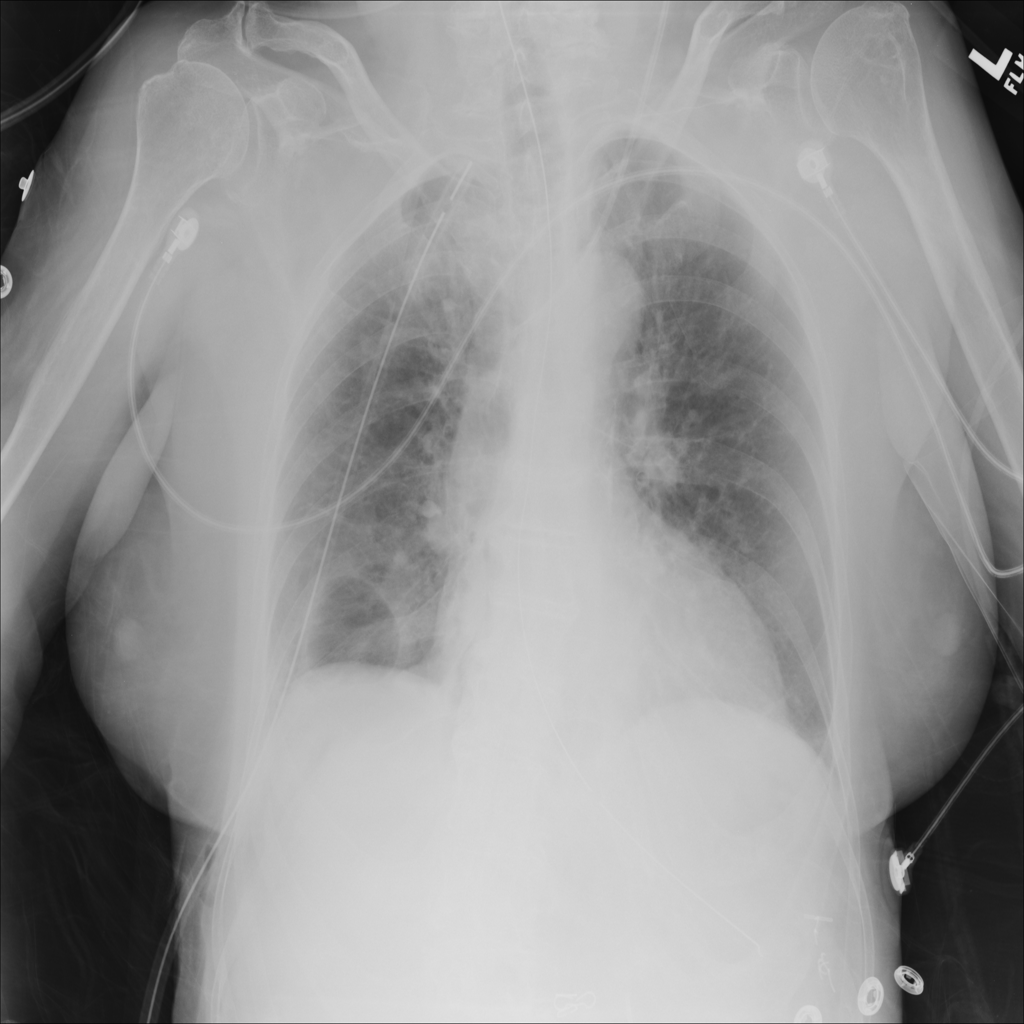}
    \end{subfigure}
    \hspace{5mm}
    \begin{subfigure}{0.3\textwidth}
        \centering
    \includegraphics[width=\textwidth]{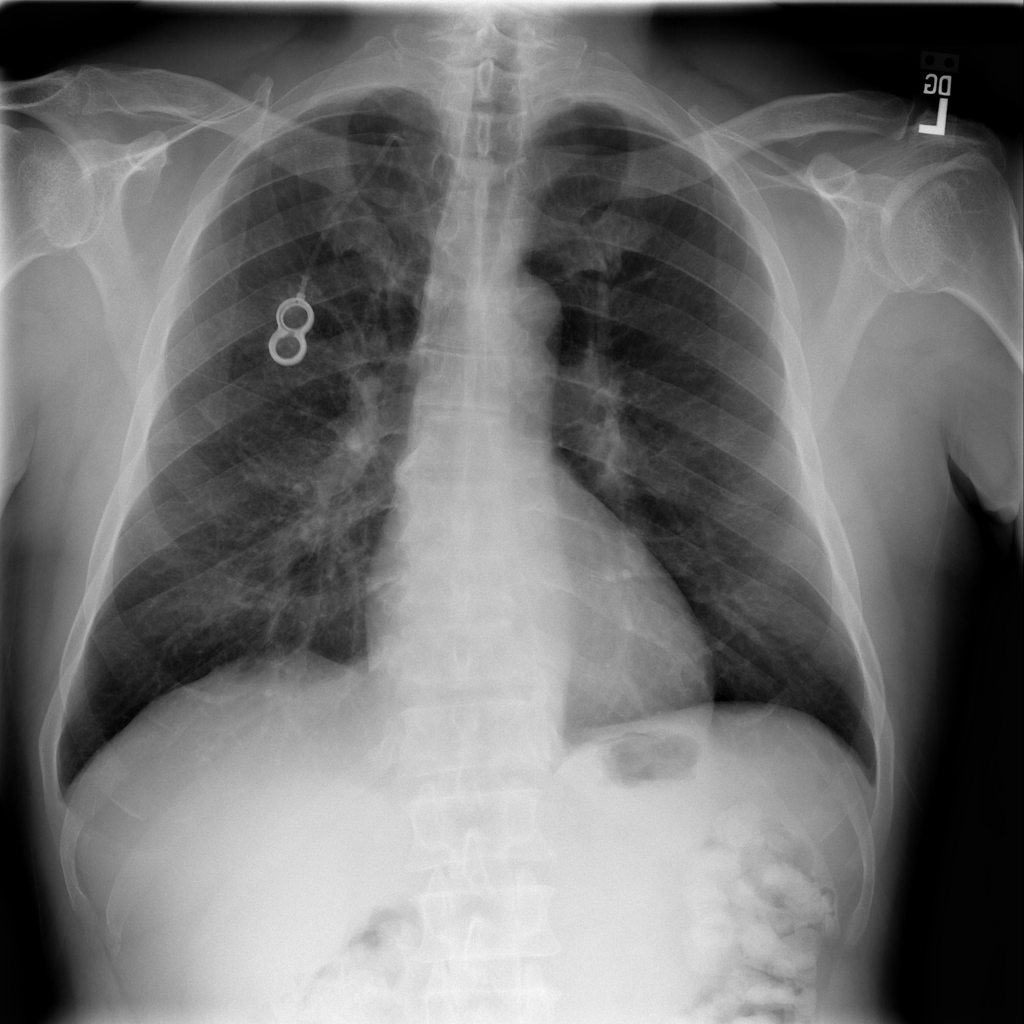}
    \end{subfigure}
    \caption{Sample images from \chestxray dataset.}
    \label{fig:example}
\end{figure}

In our work, we want the context encoder to learn how to produce patches of healthy X-ray tissue. Therefore, we first split the dataset in healthy and unhealthy images.
For the initial part of our work, we select the 61,241 healthy images from \chestxray dataset and split them into training and validation set, containing respectively 59,481 images and 1,760 images. 
These images are resized to $128 \times 128$ to be the input for the context encoder. In a first experiment, we aim to have the context encoder working with low resolutions images representing the whole chest area. For this reason and considering that abnormalities are very small with respect to the image size, we do not include unhealthy chest X-ray in the test set.

In the second part of the project, we try to predict smaller patches of healthy tissue in order to have a good-quality output at the same resolution of the source images. In order to do this, we generate a dataset of \textit{chest patches} by extracting $128 \times 128$ patches from the \chestxray dataset.
An important step when extracting patches is to make sure that we only consider the lungs area, avoiding parts from the spine, the neck and the background which would mislead the training.
For this purpose, we approximate the lungs segmentation in every image by generating two bounding boxes around them. Next, we randomly select 20 points in the bounding boxes area and crop out $128 \times 128$ patches centered in those points.
The training set for this second experiment contains 1,189,572 of these patches. For this second experiment, we use two different splits for the test set. The first one contains 880 $128 \times 128$ patches cropped from \chestxray, while the second is composed by only 33 manually annotated patches containing abnormalities.

\section{Architecture}

\begin{figure*}[h]
    \centering
    \includegraphics[width=\textwidth]{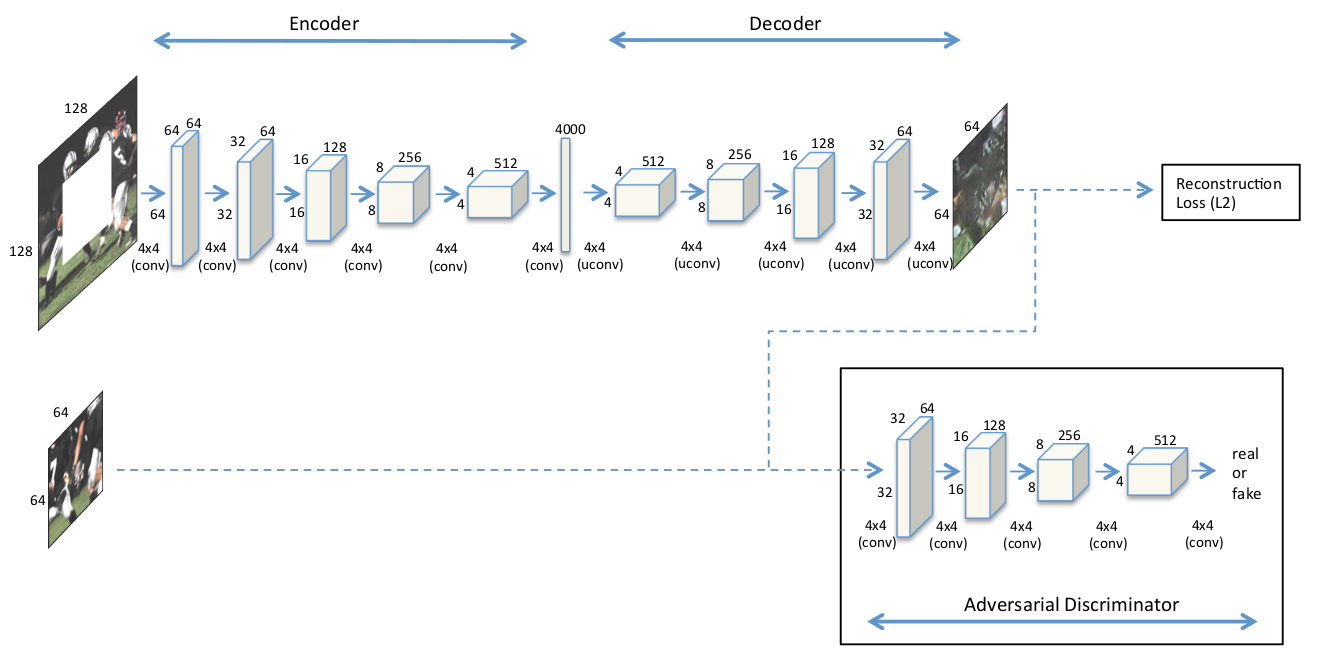}
    \caption{Figure from \cite{context-encoder}: Original context encoder architecture showing generator and discriminator FCNs}
    \label{fig:archi}
\end{figure*}

The initial architecture we choose in this project is the original context encoder from Pathak \textit{et al}. In particular, we modify an existing PyTorch implementation from Boyuan Jiang which is publicly available on GitHub\footnote{https://github.com/BoyuanJiang/context\_encoder\_pytorch}. In this Section, we explain how we tune the architecture over different experiments in order to find the configuration producing the best results. The architecture configuration of the context encoder as described in the original paper is shown in Fig. \ref{fig:archi}. We identify the generator network as the auto-encoder reconstructing the central patch given the surrounding context, and the discriminator predicting if a patch is real or fake.

\subsection{Encoder-Decoder}
\label{architecture}
The auto-encoder takes as input an image with the missing (blanked out) central patch. This \textit{context} information is then encoded through a sequence of fully convolutional layers to produce a latent feature representation of the original image. The decoder part then uses this latent representation to reconstruct the central patch. Here we present the best architecture among the ones discussed in our experiments.

\subsubsection{Encoder}
The encoder part of the architecture consists of 6 convolutional layers with kernel size 4, stride 2 and padding 1, which results in the image being halfed in size after every convolution. Each convolutional layer is followed by a batch normalization and a Leaky-ReLU activation. The input of the encoder is a grayscale image of size $128 \times 128 \times 1$, where the central $64 \times 64$ patch has been removed. After the first convolution, which brings the depth from $1$ to $128$ channels, the depth is doubled and the image size is halfed at every layer. The resulting output has shape $1 \times 1 \times 4096$. This vector is the latent feature representation of the context.

\subsubsection{Decoder}
The architecture of the decoder network is almost symmetric to the encoding one. It has 5 deconvolutional layers that take the feature vector as input and outputs a $64 \times 64 \times 1$ grayscale patch. The deconvolutions have the same kernel size, stride and padding as the convolutions in the encoder, and each layer is followed by a batch normalization and a ReLU activation. The last layer uses an hyperbolic tangent nonlinearity instead of ReLU.

\subsection{Losses and Adversarial Training}
Which loss measure should be employed to efficiently train the context encoder using back-propagation? 

\subsubsection{L2 Loss}
A loss function that is commonly used in literature is pixel-wise distance between real and reconstructed patch. Both Manhattan (L1) and Euclidean (L2) distances are valid distance measures for this purpose, with the former being more robust to outliers and the latter being more stable to small adjustments in data point values. For our work, we considered the L2 loss (reported in equation \ref{l2-loss}), since we want to penalize more generated pixels with a larger difference in intensity from the real ones, while small changes in intensity are not very important towards the quality of the final reconstruction.
\begin{equation} \label{l2-loss}
    \mathcal{L}_{L2}(x) = \Big|\Big| M \odot \Big( x - G\big( (1-M) \odot x) \big) \Big) \Big|\Big|^2 
\end{equation} 
In our notation, $x$ is the original image, $G(\cdot)$ is the generator network applied to some input image, $M$ is a binary mask matrix for the central patch and $\odot$ is the element-wise multiplication operator.

\subsubsection{Adversarial Loss}
An intrinsic issue with using L2 loss for image generation using convolutional networks is that reconstructions tend to be blurry. Among the possible solutions to this issue, \cite{adversarial-loss} propose to include an adversarial loss in the training process. This loss is computed using another convolutional network, the discriminator. This network takes as input a patch and tries to predict if it is real or reconstructed (meaning that is the output of the auto-encoder). 
The architecture of the discriminator is very similar to the encoder in the generator, with 4 convolutional $+$ batch normalization $+$ Leaky-ReLU layers. A final convolution layer results in a scalar value that is passed through a sigmoid function to output a probability value.

The training of the generator and the discriminator networks are combined, with each of them improving as the other improves. This training that is characteristic of GANs is described by the mini-max game in equation \ref{minimax-gan}. 
\begin{equation} \label{minimax-gan}
\begin{split}
\begin{aligned}
    \min_{G} \max_{D}V(G, D) = \min_{G} \max_{D} \mathbb{E}_{x \in \mathcal{X}} \Big[& \log \big(D(M \odot x)\big) 
    + \log \big( 1-D( G((1-M) \odot x))\big) \Big]
\end{aligned}
\end{split}
\end{equation}

The adversarial loss for the generator part derived from the mini-max game is shown in Eq.  \ref{adversarial-loss}. In this case, we want the generator to trick the discriminator into categorizing reconstructed images as original ones (meaning that they get score $1$). Since the evaluation of original images does not influence the training of the generator, that part of the adversarial loss actually omitted in the implementation.
\begin{equation} \label{adversarial-loss}
\begin{split}
\begin{aligned}
    \mathcal{L}_{adv} = \max_D \mathbb{E}_{x \in \mathcal{X}} \Big[& \log \big(D(M \odot x)\big) 
    + \log \big( 1-D( G((1-M) \odot x))\big) \Big] 
\end{aligned}
\end{split}
\end{equation}

From the mini-max game we can also derive the adversarial loss for the discriminator, which is defined as the binary cross entropy between the true and the predicted distributions of real and reconstructed images. In the optimal case, the discriminator outputs $1$ as the probability of original images to be true, and $0$ as the score for reconstructed images to be true.

In an alternate manner for every training batch, one of the two networks (G and D) is kept fixed and the other is updated. When a convergence situation is reached (in an ideal case), the generator can perfectly reconstruct the original image and the discriminator is not able anymore to discern fake from real, assigning probability 0.5 to every image (original or reconstructed). In this situation where both $D(M \odot x) = \frac{1}{2}$ and $D\big(G((1-M) \odot x) = \frac{1}{2}$, the optimal value for the mini-max game in Eq. \ref{minimax-gan} converges to the value $ V(G, D) = -\log 4$. Note that this loss indicates convergence in the training because the discriminator is no more able to teach the generator how to improve, but it does not ensure that the generator can generate good-quality reconstructions. In the following sections we describe some ways to balance generator and discriminator in order to avoid convergence to sub-optimal performance.

\subsubsection{Joint Loss}
The final joint loss that we use to train our generator is a weighted average of the L2 and adversarial losses, as shown in Equation \ref{joint-loss}.
\begin{equation} \label{joint-loss}
    \mathcal{L} = \lambda_{L2} \mathcal{L}_{L2} + \lambda_{adv} \mathcal{L}_{adv}
\end{equation}
We experimented with different weighting for the combination and found the best results with $\lambda_{L2} = 0.998$ and $\lambda_{adv} = 0.002$, which are very similar to the values chosen in the original context encoder paper. Increasing the weight for the adversarial loss generates a realistic patch that, however, does not semantically connect with the surrounding context. A larger weight for the L2 loss would result in the original issue of a blurry reconstruction.

%


During different experiment, in addition to visually inspecting the reconstruction, we analyze the training curves for losses and predictions to understand how the context-encoder is learning the task. In our final model we notice that in the initial part of the training the score of the mini-max game $ V(G, D)$ fluctuates around $- \log 4$. At the same time, the average probability for reconstructed images to be classified as original images is slightly lower than the probability of source images to be classified as original, with both scores being close to $0.5$. This means that the discriminator and generator are well balanced. After around 40 epochs, however, only the discriminator network continues improving, while the generator stops. This behavior is expected and caused by the constraint enforced by the L2 loss. In more details, the adversarial signal from the discriminator would cause the L2 loss to increase too much and, as a consequence, the joint loss combining L2 and adversarial would start increasing.

\subsection{Balancing Generator and Discriminator}
\label{balancing}
Although the definition of adversarial training is mathematically founded, in real case scenarios it is not easy to obtain a good convergence of the model, especially when the discriminator and generator components are not balanced enough. We have seen that when generator and discriminator are balanced, the mini-max game reaches the value $ V(G, D) = - \log4$. If, instead, this score is lower it means that the discriminator is too powerful with respect to the generator, which, in turn will not be able to learn from the signal received from the discriminator. On the other hand, when the generator is better than the discriminator, the generator does not have any accurate adversarial signal to improve upon. In the optimal scenario, both networks improve gradually together, and only after enough epochs convergence is reached.

To solve this balancing issue, we try different ways to tune the architecture.
First of all, we can change the number, depth and stride of convolutional layers in both networks. In addition, we can freeze one of the two parts during training so that the other one runs over more training samples. This can be done both epoch-wise or iteration-wise. We also try to train the generator network only with L2 loss in the beginning, in order to have much better initial reconstructions on top of which to start the adversarial training.
A crucial component of the architecture is the size of the bottleneck for the latent representation of the image in the context encoder. Changing this value enables a larger or more compact feature representation, but also heavily influences the number of weights in the architecture. Other parameters we considered are the values for Adam optimizer, with a learning rate starting at $0.0002$ and betas $\{0.5; 0.999\}$. These values are picked after running some experiment applying grid search on hyper-parameters.

\subsection{Achieving Local and Global consistency}
After few initial experiments, we noticed, by visually inspecting reconstructed images, that the most evident artifact was the margin between the generated patch and the original context. 

A solution to this problem was proposed in the original Context Encoder paper by Pathak et al. The simple modification they suggest consists in weighting the L2 loss in a way that areas around the margin are more relevant than the central ones. Indeed, there has to be a constraint on the continuity in shapes and brightness between the old and the reconstructed tissue. In our case, we decided to multiply by $10$ the loss for a $4$-pixel-wide region at the borders of the central patch.
This solution was proven to be very effective and was kept in all latter experiments.


\section{Experiments and Results}
\subsection{Inpainting on the Whole Image}
As the first part of the project, we focused on the whole content of the X-ray scan. In particular, given the whole image we crop out the central part corresponding to the chest area and try to reconstruct it using our model. As a first pre-processing step for this experiment we resize the original $1024 \times 1024$ images to lower resolution $128 \times 128$ ones. This $64\times$ size compression is necessary due to the large complications in training the network for high resolution images. Even at this low resolution, at least 10 Gb of GPU memory on a cluster are necessary to store the network weights during training.

\begin{figure}[h]
    \centering
    \begin{subfigure}{0.45\textwidth}
        \centering
        \includegraphics[width=\textwidth]{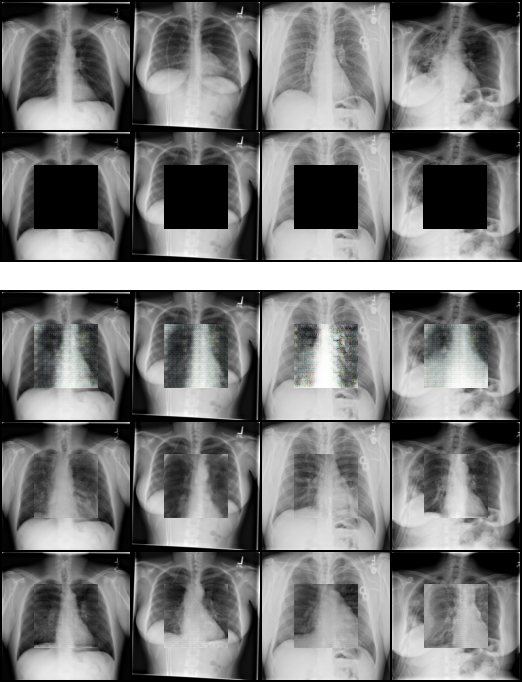}
        \caption{\textbf{Initial results without L2 loss.} First row: 4 random original chest X-rays. Second row: masked images. Third row: reconstructions with the original configuration of the context-encoder. Fourth row: our reconstructions freezing the generator. Fifth row: our reconstructions increasing the depth in convolutions of the discriminator.}
        \label{fig:whole-adv}
    \end{subfigure}
    \hspace{5mm}
    \begin{subfigure}{0.45\textwidth}
        \centering
        \includegraphics[width=\textwidth]{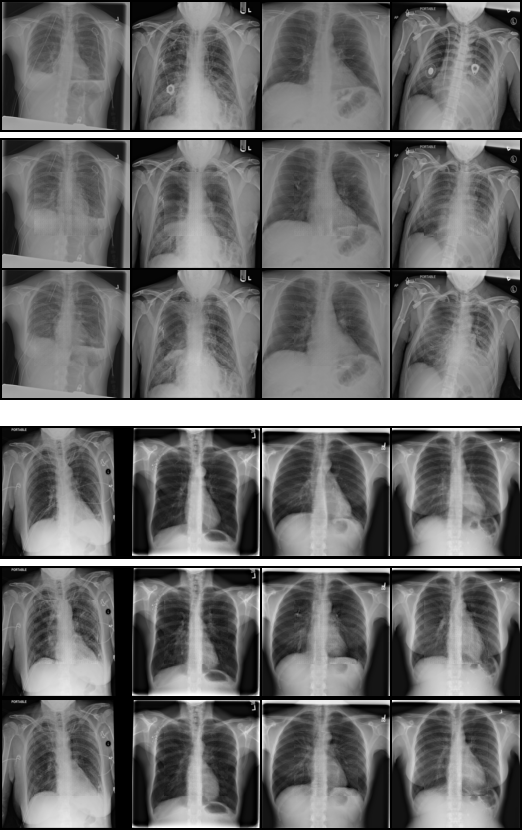}
        \caption{\textbf{Reintroducing the L2 loss.} The original images in the first row are compared with the ones reconstructed freezing the discriminator (second row) and increasing the depth in convolutions (third row). The same structure is used in the second block of three rows. Both techniques result in accurate inpaintings.}
        \label{fig:whole-final}
    \end{subfigure}
    \caption{Reconstructing the whole image with only the adversarial loss (Subfig. a) and with the joint loss (Subfig. b).}
\end{figure}

Since the adversarial loss for the generator only accounts for a little amount towards the total loss, it is hard to visualize when the adversarial training using the discriminator is working effectively. Of course, when the discriminator is not learning at all the reconstructions are very blurry due to only considering the L2 loss. To understand and improve the adversarial part of our model, we initially remove the L2 loss from our joint loss. In this way, the model only relies on the signal from the discriminator to learn the task.
We tried different ways to tune and enhance the generator-discriminator pair of networks. Among many results, we compare the reconstructions using the original context-encoder architecture with two of the best modified architectures and the true chest X-ray. As we show in Fig. \ref{fig:whole-adv} with some sample images, the initial configuration of the context-encoder is not particularly good at reconstructing the inpainted patch. For example, the ribs are not present and the whole area appears still blurry. (Remind that the discriminator network only evaluates then central patch, thus in this case the gap between the patch and the surrounding context is evident. This smooth transition is enforced later on when reintroducing the L2 loss.)
Looking at the loss values in this configuration, we notice that the discriminator network is not very good in understanding which images are the real ones. To solve this issue, we applied the techniques previously explained in \ref{balancing}.
The second group of reconstructions are obtained by freezing the generator network once every two iterations in order to train the discriminator more frequently. It is clear that the central patches are more realistic than in the basic version. As a third experiment, we show how results can also be improved by increasing the capacity of the discriminator network. In this case, the depth of every convolutional layer is doubled with respect to the original configuration, starting at $128$ channels instead of $64$ for the first convolution.
\footnote{All of the results in this experiment are obtained after 20 epochs of training on the whole dataset, which takes around 15 hours on a Nvidia GTX 650M GPU.}

    

Once the adversarial training has been fine-tuned, we can reintroduce the L2 loss which helps with obtaining a smoother reconstruction of the whole image, where the shapes in the central patch blend in with the surrounding context.
In Fig. \ref{fig:whole-final} we show some reconstructions from the test set using the two models described before after 20 epochs of training. The X-rays generated from our model appear very realistic and it is hard to distinguish them from the original images. It is interesting to notice how some abnormal circular artifacts present in the second and fourth images are removed in the reconstructed patch. Also, the reconstructions are accurate even when the X-ray are rotated (fourth image), with the generated ribs perfectly aligned to the ones in the external region.


\subsection{Inpainting a High-resolution Patch}

\begin{figure}[h]
    \centering
    \begin{subfigure}{0.45\textwidth}
        \centering
        \includegraphics[width=\textwidth]{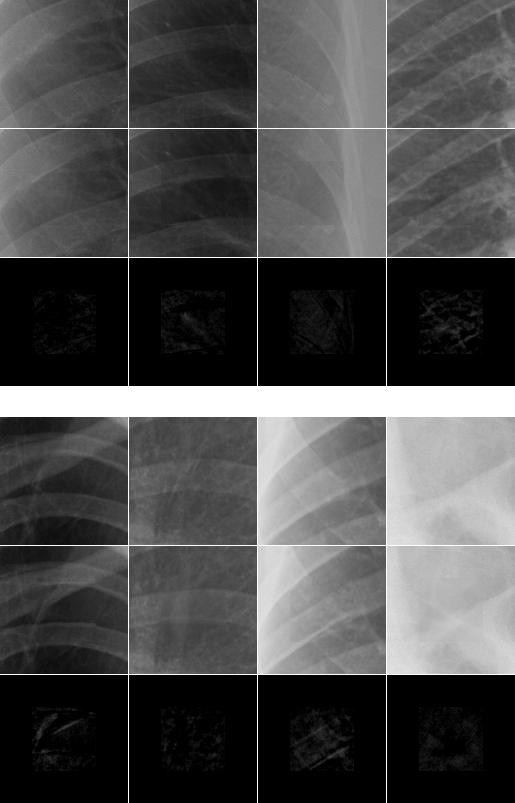}
        \caption{\textbf{Results on high-resolution healthy patches.} Reconstructing tissue from healthy patches. The first and fourth row show the original images, while the second and fifth are the reconstructed versions and the third and sixth show the pixel-wise absolute difference between the pairs above. Since no abnormalities are present and the reconstruction are very accurate, these images are almost completely black.}
    \label{fig:healthy}
    \end{subfigure}
    \hspace{5mm}
    \begin{subfigure}{0.45\textwidth}
        \centering
        \includegraphics[width=\textwidth]{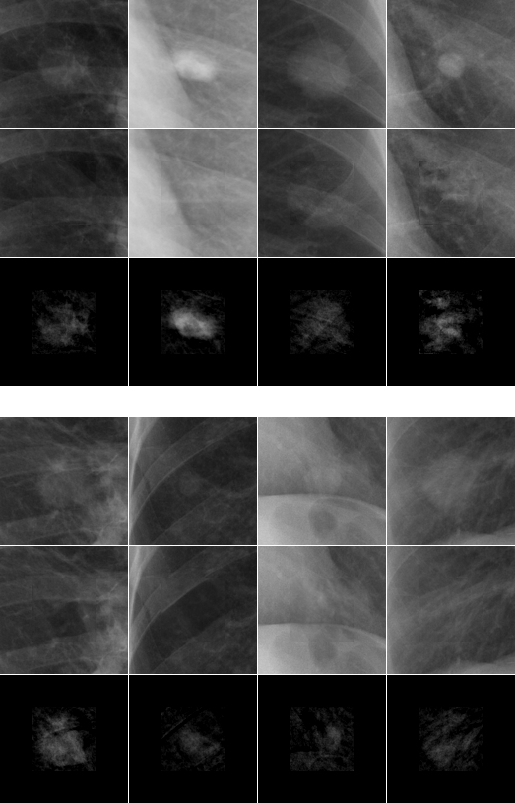}
        \caption{\textbf{Results on high-resolution unhealthy patches.} Abnormalities in the center of the patches in rows 1 and 4 are removed in the reconstructions generated in rows 2 and 5. These sets of images show the healthy reconstructed version of the original diseased X-ray patch. In row 3 and 6 we show the difference between the original and reconstructed patches, where the abnormality shapes are now evident.}
    \label{fig:unhealthy}
    \end{subfigure}
    \caption{Reconstructing high-resolution patches of healthy or unhealthy tissue.}
\end{figure}

Since the abnormalities in chest X-rays are often very small in comparison to the full size of the image, it is obvious that down-scaling by $64\times$ the image results in losing a lot of important information when working towards a following detection task. Nevertheless, with the previous experiments we understood how to tune the adversarial training to efficiently work with $128 \times 128$ images. With this in mind, we start working with full-resolution smaller patches from the chest images. Instead of teaching our model how to reconstruct the whole chest area, we want it to reconstruct smaller portions of tissues from different parts of the chest. Since the original images also show different regions such as spine, neck and background, this patches should not be considered for training as they would be misleading.
For this purpose we use the dataset generated by cropping patches from the lungs areas, as explained previously in section \ref{dataset}.
After trying different configurations for our model, we chose as the best final architecture the one described in section \ref{architecture}.
In Fig. \ref{fig:healthy} we show the results obtained after training our model for 90 epochs.
In the first two epochs we only train the generator using L2 loss, then we train the discriminator for 4 epochs, and finally we train discriminator and generator combined with the joint loss.
Once again, the reconstructions appear very realistic, with continuity in shapes and texture between the central patch and the surrounding region and a high similarity between the content of the reconstructed and the original patches. For every pair of real and reconstructed images we also generate a third image highlighting their differences. In particular, we take the absolute value of the pixel-wise difference between the two images and double it to make the changes more evident. Analyzing these images, we notice that there are very little differences and, if present, they appear in form of random noise without a recognizable shape. This tells us that, up to small modifications, our reconstruction is true to the original.

As a second test set of images for this experiment, we also consider chest X-rays showing some diseased tissue. Thanks to an expert annotator, we select 33 patches showing some type of abnormalities in the central region. Those images are fed to our model, which removes the diseased central areas trying to reconstruct an healthy version of them, given the information present in the surrounding healthy regions. In Fig. \ref{fig:unhealthy} we show that our model is able to generate meaningful inpaintings in most of the cases, where the abnormality in the original X-ray is not present anymore. The changes introduced in the reconstructed healthy versions of the patches can be noticed when looking at the images showing the pixel-wise difference. In contrast to the healthy test set, here we see a precise shape in the central patch which reflects the abnormalities present in the original unhealthy X-ray.

\subsection{Qualitative and Quantitative Evaluation}

Other than a visual inspection to validate the performance of our model, we also want some more accurate and reliable evaluation of our model's performance. As quantitative measures for the reconstructions we employed the PSNR, MSE and SSIM metrics, which are often chosen in literature to describe the quality of a reconstructed or compressed image in comparison to the original. However, we want to point out that in our task we do not want the reconstruction to be a copy of the source image. Indeed, many different patches could correctly fill the missing region, and each of them would represent a meaningful reconstruction. On the other hand, it is true that each of those patches should be similar to a certain extent to the original ones. Confirming our intuition, the average PSNR (Peak Signal-to-Noise Ratio) value on the healthy test set is $30.88$ dB with a standard deviation of $3.61$ dB, which shows that the images are similar only to some extent. For comparison, we also compute the average PSNR on the unhealthy labeled test set. In this case the score drops to $27.06 \pm 3.25$ dB, showing that the healthy reconstructions generated from our model have larger differences with the unhealthy original versions.
As a second measure, we consider the MSE (Mean Squared Error) to describe how much the pixel intensities deviate between the original and reconstructed patch. The average MSE between real and reconstructed patches in the healthy test set is $79.20 \pm 129.16$, while the score we obtain on the unhealthy test set is $164.82 \pm 122.95$, confirming that healthy reconstructions are more similar to the original patches.
Finally, we include the SSIM (Structural Similarity) index in our quantitative evaluation. This perception-based measure is different to absolute errors mentioned previously as it describes the perceived change in structural information. From a practical point of view, this measure is commonly used in predicting the perceived quality of digital television and cinematic pictures, as well as other kinds of digital images and videos.
The average SSIM signal resulting on our healthy test set is $0.81 \pm 0.09$, while the score on the unhealthy test set is $0.76 \pm 0.08$. Considering that the SSIM signal for identical images is $1$, these scores confirm the observations from the previous visual inspection, where reconstructions on the healthy test set are more similar to the originals then the one on unhealthy patches, since the abnormalities are removed.

\begin{table}[t]
\begin{center}
\begin{small}
\begin{sc}
\begin{tabular}{lll}
\toprule
Measure & Healthy test set & Unhealthy test set \\ \midrule
PSNR & $30.88 \pm 3.61$        & $27.06 \pm 3.25$            \\ 
MSE & $79.20 \pm 129.16$        & $164.82 \pm 122.95$          \\ 
SSIM & $0.81 \pm 0.09$       & $0.76 \pm 0.08$           \\ 
\bottomrule
\end{tabular}
\end{sc}
\end{small}
\end{center}
\caption{Scores for the reconstructions averaged on healthy and unhealthy test sets.}
\label{tab:measures}
\end{table}

Since this quantitative measure does not provide an absolute description of our results, we also performed a qualitative evaluation of the reconstructions. To our knowledge, no other projects worked at inpainting on the \chestxray dataset. Therefore, we decided to validate the results by conducting a two-alternative forced choice (2AFC) study.
In this test an expert is asked to find, from a pair of $1024 \times 1024$ images, which of them is the original one and which the one with an inpainted patch generated from our model.
The accuracy of the observer in this case study was $59.45 \%$, that means in almost half of the cases they were unable to discriminate the reconstructed version from the original one.

\section{Conclusion and Future Work}
This paper demonstrated how the context encoder architecture under adversarial training can effectively be used for inpainting in chest X-ray images. Having a model which is able to inpaint healthy tissues given the surrounding regions opens up the possibility to solve in a new way crucial tasks like automated abnormality detection. Our architecture is able to compress all the useful information from the surrounding context in a latent representation, which is used to reconstruct an healthy version of the target patch. We have proven the quality of the reconstructions with a two-alternative choice observer study.

While we have chosen the context encoder architecture to accomplish the inpainting task, there is a variety of architectures under the GAN framework which could also be employed for the same goal. For example, a recently published paper \cite{partial-conv} proposes the use of partial convolutions for inpainting, where convolution layers are masked and renormalized to be conditioned on only valid pixels. Also, we would like to experiment with the approach proposed by \cite{global-and-local} in order to improve the coherence and shape continuity in the reconstructions. In this approach, the discriminator network combines a local path (similar to the discriminator in our work) which focuses on detecting fine-grained artifacts in the central patch, and a global path which looks for a high-level coherence in the final image, in particular in the critique region between the context area and the inpainted patch. 
Furthermore, building on the results shown in this work, a classifier could be trained on top of the images showing the pixel-wise difference between real and reconstructed X-rays to automatically understand if and which diseases appear in different patches, or to compute regression over position and dimension of the abnormality.


\bibliographystyle{unsrt}
\bibliography{project_refs}

\end{document}